\pdfoutput=1
\documentclass[11pt]{article}

\usepackage[final]{acl}

\usepackage{times}
\usepackage{latexsym}

\usepackage[T1]{fontenc}
\usepackage[utf8]{inputenc}

\usepackage[nopatch=footnote]{microtype}

\usepackage{inconsolata}

\usepackage{amsmath}
\usepackage{amssymb}
\usepackage{mathtools}
\usepackage{amsthm}

\usepackage[capitalize,noabbrev]{cleveref}

\usepackage{yhmath}

\usepackage{multirow}
\usepackage{soul}
\usepackage{mathtools}
\usepackage{amsmath,amsfonts,amsthm}
\usepackage{algorithm, algpseudocode}
\usepackage{graphicx}
\usepackage{float}
\usepackage{lipsum}
\usepackage{textcomp}
\usepackage{xcolor}
\usepackage{soul}
\usepackage{placeins}
\usepackage{hhline}
\usepackage{tabularx}
\usepackage{enumitem}
\usepackage{setspace}
\usepackage{booktabs}
\usepackage{adjustbox}
\usepackage[normalem]{ulem}
\usepackage{indentfirst}
\usepackage{titlesec}
\usepackage{wrapfig}
\usepackage{array}
\usepackage{colortbl}
\usepackage{subcaption}

\graphicspath{{figures/}}

\theoremstyle{plain}
\newtheorem{theorem}{Theorem}[section]

\newtheorem{corollary}[theorem]{Corollary}
\theoremstyle{definition}

\theoremstyle{remark}

\title{FedSpaLLM: Federated Pruning of Large Language Models}

  \author{Guangji Bai$^{1,2}$\thanks{This work was done during Guangji Bai's internship at Argonne National Laboratory.} \quad Yijiang Li$^{1}$ \quad Zilinghan Li$^{1}$ \quad Liang Zhao$^{2}$ \quad Kibaek Kim$^{1}$\thanks{Corresponding Author}  \\
  $^{1}$Argonne National Laboratory \quad $^{2}$Emory University  \\
  \texttt{\{guangji.bai,liang.zhao\}@emory.edu} \\  \texttt{zl52@illinois.edu}, \quad  \texttt{\{yijiang.li,kimk\}@anl.gov}
}

\begin{document}
\maketitle

\begin{abstract}
Large Language Models (LLMs) achieve state-of-the-art performance but are challenging to deploy due to their high computational and storage demands. Pruning can reduce model size, yet existing methods assume public access to calibration data, which is impractical for privacy-sensitive applications.
To address the challenge of pruning LLMs in privacy-preserving settings, we propose \textbf{FedSpaLLM}, the first federated learning framework designed specifically for pruning LLMs. FedSpaLLM enables clients to locally prune their models based on private data while accounting for system heterogeneity and maintaining communication efficiency. Our framework introduces several key innovations: (1) a novel $\ell_0$-norm aggregation function that ensures only non-zero weights are averaged across clients, preserving important model parameters; (2) an adaptive mask expansion technique that meets global sparsity targets while accommodating client-specific pruning decisions; and (3) a layer sampling strategy that reduces communication overhead and personalizes the pruning process based on client resources.
Extensive experiments show that FedSpaLLM improves pruning performance in diverse federated settings. The source code can be found at~\url{https://github.com/BaiTheBest/FedSpaLLM}.
\end{abstract}

\section{Introduction}
\label{sec:introduction}

Large Language Models (LLMs) such as GPT~\cite{openai2023gpt} and LlaMA~\cite{touvron2023llama} have recently transformed the field of Natural Language Processing (NLP) due to their ability to perform exceptionally well across a variety of complex language benchmarks. However, the increasing scale of these models, which can contain billions of parameters, also brings significant computational and storage costs. The high memory and inference costs make it challenging to deploy LLMs in real-world applications where resources are constrained~\cite{bai2024beyond}. Consequently, there has been an increasing interest in \emph{model compression} techniques such as pruning, quantization, and knowledge distillation, which aim to reduce the computational load while maintaining model performance~\cite{zhu2023survey}. Among these techniques, \emph{pruning} has emerged as a highly effective approach for reducing the size and complexity of LLMs by introducing sparsity into the models.

Despite the recent success of LLM pruning methods, existing approaches predominantly assume that the calibration data used for pruning is publicly available~\cite{frantar2023massive,sun2023simple}. However, in many real-world scenarios, especially when dealing with sensitive applications like medical agents or financial systems, the data used for pruning might be private and cannot be shared openly~\cite{ren2024advances}. On the other hand, Federated Learning (FL), a distributed machine learning technique that enables multiple clients to collaboratively train models without sharing their private data, has gained significant popularity in traditional machine learning~\cite{zhang2021survey}. However, most works on LLMs in FL settings have focused on fine-tuning. Due to the intrinsic differences between fine-tuning and pruning, existing FL-based fine-tuning methods cannot handle the problem of pruning LLMs with private data.

To address the challenges posed by pruning LLMs in federated settings, where private data cannot be shared and heterogeneity exists among clients, we propose a novel method called \textbf{FedSpaLLM} (Federated Sparse LLM). FedSpaLLM is the first framework that allows pruning LLMs under a federated learning setting with resource heterogeneity. Our method allows each client to prune its local model based on its data while maintaining privacy and accommodating diverse computational resources.

Our contributions are summarized as follows:
\begin{itemize}[leftmargin=*, itemsep=-2pt]
\vspace{-2mm}
    \item \textbf{Federated pruning framework for LLMs}: We present the first framework for pruning LLMs in FL, allowing collaborative pruning with private local data.
    \item \textbf{$\ell_0$-norm aggregation}: We introduce a novel aggregation function that preserves important weights by averaging only non-zero elements across client models.
    \item \textbf{Adaptive mask expansion}: We propose a mask expansion technique to meet global sparsity targets while accounting for client-specific pruning.
    \item \textbf{Layer sampling}: We develop a resource-aware layer sampling strategy, enabling personalized pruning and reducing communication costs.
    \item \textbf{Extensive evaluation}: We conduct comprehensive experiments, showing that FedSpaLLM improves both pruning efficiency and model performance in heterogeneous federated environments.
\end{itemize}

\section{Related Work}
\label{sec:related work}

\noindent\textbf{Pruning of LLMs.}
\emph{Pruning} regained prominence in the late 2010s for reducing inference costs~\cite{han2015deep}. LLM pruning can be categorized into \emph{structured} and \emph{unstructured} pruning.

Unstructured pruning removes individual parameters without regard to model structure, often using thresholds to nullify smaller weights. SparseGPT~\cite{frantar2023massive} achieves up to 60\% parameter reduction in LLMs with minimal performance loss. Wanda~\cite{sun2023simple} introduces a pruning criterion based on both weight magnitude and activations, particularly effective in linear layers. DynaTran~\cite{tuli2023acceltran} dynamically prunes activations at runtime, enhanced by a custom ASIC architecture.

Structured pruning removes groups of parameters such as filters or attention heads. LLM-Pruner~\cite{ma2023llm} combines first-order data and Hessian information for structured pruning, while LoSparse~\cite{li2023losparse} uses low-rank and sparse approximations to balance pruning and model expressiveness. Structured pruning of hidden dimensions, as shown by~\cite{tao2023structured}, extends to embeddings and attention heads. ZipLM~\cite{kurtic2023ziplm} optimizes structured compression for accuracy and hardware efficiency.

\noindent\textbf{Federated Learning with LLMs.}
FL on LLMs has primarily focused on LLM \emph{fine-tuning} and has gained attention for enabling private and efficient model updates. FedPrompt \citep{zhao2023fedprompt} introduces prompt-tuning in FL, reducing communication costs by updating only soft prompts. \citet{fang2024fedpipe} and \citet{li2024secure} leverage low-rank adapters for parameter-efficient fine-tuning, improving training speed and accuracy. FedNLP \citep{lin2022fednlp} provides a benchmarking framework for evaluating FL methods on NLP tasks. FedAdapter \citep{cai2023fedadapter} uses adapters to accelerate model convergence in federated settings. FeDeRA \citep{yan2024federa} employs singular value decomposition to further improve LoRA-based fine-tuning efficiency. C2A \citep{kim2023c2a} introduces a hypernetwork-based framework for generating client-specific adapters to handle client heterogeneity. FedBPT \citep{sun2024fedbpt} enables efficient prompt-tuning with a gradient-free approach, reducing memory and communication costs. PrE-Text \citep{hou2024pre} generates differentially private synthetic data to enable central training, reducing on-device computation.

\noindent\textbf{Federated Pruning on DNNs.}
Model pruning in FL improves efficiency by reducing communication and computation costs. FedP3 \citep{yi2024fedp3} and HeteroFL \citep{diao2021heterofl} address client model heterogeneity, enabling smaller, personalized models. FedTiny \citep{huang2023fedtiny} and PruneFL \citep{jiang2022prunefl} implement progressive pruning to fit models within resource constraints. FedPrune \citep{munir2021fedprune} improves performance by pruning global models based on client capabilities, while Complement Sparsification \citep{jiang2023complement} reduces communication overhead using sparsity techniques. However, all the work above only applies to smaller DNNs, and cannot trivially scale to massive LLMs.

\noindent\textbf{Federated Distillation.}
In addition to model pruning, knowledge distillation (KD) has been explored as a resource-efficient approach to training NNs in edge environments~\cite{li2024federated}. In this paradigm, a smaller proxy model is used to facilitate learning, either by mimicking the behavior of a larger model or by aggregating knowledge from multiple clients. For instance, FedBiOT~\cite{wu2024fedbiot} introduces a method for locally fine-tuning LLMs without requiring full-model transmission, allowing clients to distill and learn from a proxy model. Similarly, DaFKD~\cite{wang2023dafkd}, a domain-aware FL knowledge distillation framework, incorporates domain-specific adaptations to improve the generalization of models across heterogeneous clients.

Despite the merits of federated distillation, our proposed approach offers several advantages. Proxy models and FL distillation methods require training additional models, increasing computational overhead. Additionally, they assume an accessible public dataset for generating the soft labels, which usually does not hold in practice. Our framework directly prunes the global LLM without auxiliary models and any additional public data. It achieves target sparsity across heterogeneous clients while maintaining privacy through local pruning and aggregation.

\section{Preliminaries}
\label{sec:preliminary}

\subsection{Pruning of LLMs}
 
In this work, we focus on \emph{unstructured} pruning,~\cite{frantar2023massive,sun2023simple,bai2024sparsellm} which typically utilizes local pruning. Local pruning circumvents the memory issue mentioned above by dividing the full model compression into sub-problems for each layer and constructing a \emph{local loss} to measure the $\ell_2$-error between the outputs of the uncompressed and compressed layers. Hence, the local pruning can be formulated by:
\begin{equation}
\text{min}_{\mathbf{M}_{\ell}, \widehat{\mathbf{W}}_{\ell}} \Vert \mathbf{W}_{\ell} \cdot \mathbf{X}_{\ell} - (\mathbf{M}_{\ell} \odot \widehat{\mathbf{W}}_{\ell}) \cdot \mathbf{X}_{\ell} \Vert_2^2,
\label{eq: layer wise pruning}
\end{equation}
where $\odot$ denotes element-wise multiplication.
% The equation above represents the core mathematical formulation used in local pruning.
Although smaller than the global pruning, the local pruning still needs to optimize both the mask $\mathbf{M}_{\ell}$ and the remaining weights $\widehat{\mathbf{W}}_{\ell}$ and thus remains NP-hard.
% A key aspect of the local pruning problem in Eq.~\ref{eq: layer wise pruning} is that both the mask $\mathbf{M}_{\ell}$ and the remaining weights $\widehat{\mathbf{W}}_{\ell}$ are optimized jointly, hence the problem remains NP-hard. 
Therefore, exactly solving it for larger layers is unrealistic, leading all existing methods to resort to approximations.

\paragraph{Mask Selection \& Weight Reconstruction.} 
A particularly popular approach is to separate the problem into \textit{mask selection} and \textit{weight reconstruction}~\cite{hubara2021accelerated,kwon2022fast}. Concretely, this means first choosing a pruning mask $\mathbf{M}$ according to some salient criterion, like the weight magnitude~\cite{zhu2017prune}, and then optimizing the remaining unpruned weights while keeping the mask unchanged. Importantly, once the mask is fixed, Eq.~\ref{eq: layer wise pruning} turns into a \textit{linear regression} problem that can be easily optimized.

\subsection{Background of Federated Learning}

FL is a distributed machine learning paradigm designed to enable collaborative training of a single global model without exchanging private data between clients~\cite{gu2023dynamic}. In this setup, multiple clients, each with their own local dataset, train their local models independently. The central server coordinates the process by sending the global model to a selected group of clients. These clients then perform local optimization using their respective datasets. After local training, the updated models are sent back to the server, where they are aggregated to update the global model.

In each communication round, the server aggregates the updates from multiple clients to refine the global model. Mathematically, let $\mathcal{D}_i$ denote the local dataset of the $i$-th client, and $\theta$ represent the global model parameters. The process of updating the global model is given by:
\begin{equation}
\tilde{\theta} \in \arg\min \sum\nolimits_{i=1}^{K} \alpha_i \cdot \mathcal{L}(\mathcal{D}_i; \theta),
\label{eq:fl_definition}
\end{equation}
where $\mathcal{L}(\mathcal{D}_i; \theta)$ is the local objective function computed over the dataset $\mathcal{D}_i$ with the model parameters $\theta$. The factor $\alpha_i$ is typically proportional to the size of client $i$'s dataset, i.e., $\alpha_i = \frac{|\mathcal{D}_i|}{\sum_{i} |\mathcal{D}_i|}$, and $K$ denotes the total number of participating clients. 

There exist various methods for aggregating client updates, with FedAvg~\cite{mcmahan2017communication} being the most widely adopted due to its simplicity and effectiveness in a wide range of FL applications. This method aggregates the client updates weighted by their respective dataset sizes, providing a robust way to train models in data-sensitive environments.

\begin{figure*}[t!]
  \begin{center}
    \includegraphics[width=1\textwidth]{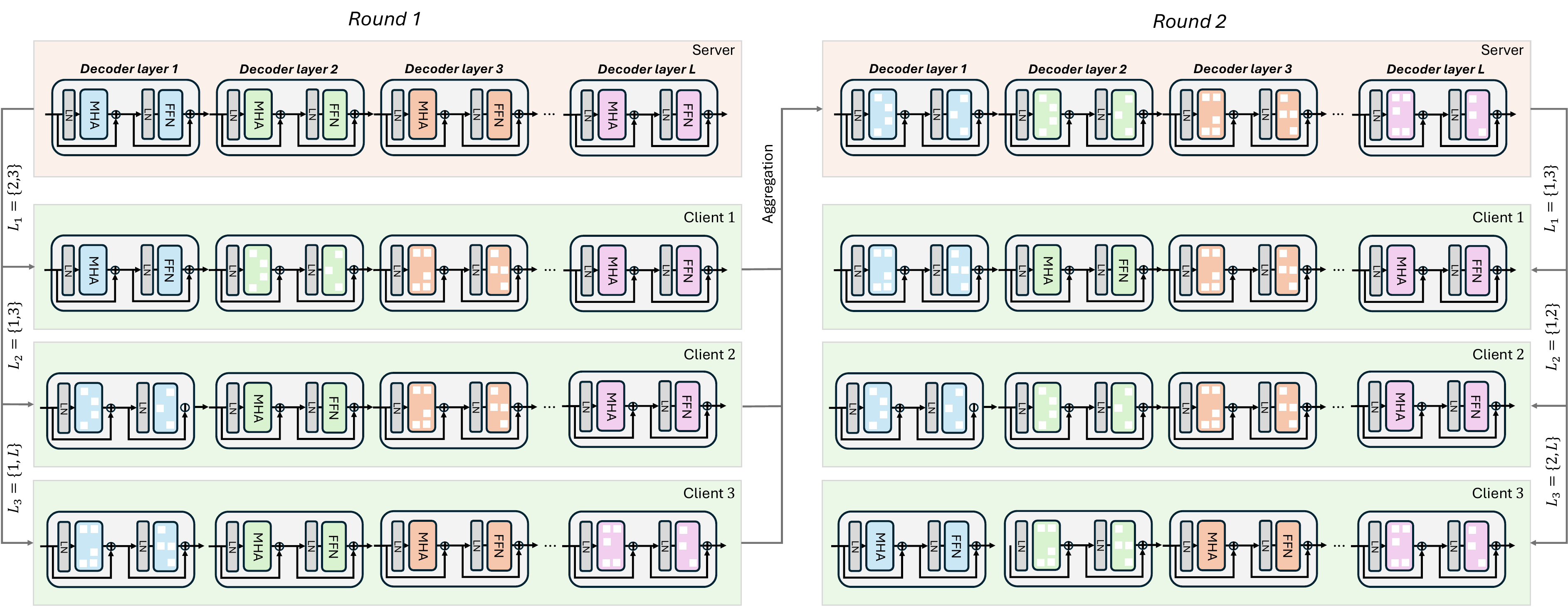}
  \end{center}
  \vspace{-2mm}
  \caption{Visualization of the proposed \textbf{FedSpaLLM} framework. Instead of transmitting the full model at each communication round, the server samples a subset of layers based on each client’s computational resources. Clients prune only the sampled layers and retain the rest from their cached pre-trained dense model. After local pruning, clients only send their pruned layers to the server, which aggregates the pruned layers using a novel $\ell_0$-norm aggregation function that averages only the non-zero parameters. This approach ensures that important weights are preserved while reducing communication overhead. The layer sampling strategy enables personalized pruning tailored to client heterogeneity, reducing resource usage without compromising overall model performance.}
  \label{fig: framework}
  \vspace{-4mm}
\end{figure*}

\section{Proposed Method}
\label{sec:proposed_method}

In this section, we present FedSpaLLM, our novel framework for federated pruning of large language models (LLMs). The proposed method is designed to address the challenges of pruning in federated learning settings, specifically targeting communication efficiency and system heterogeneity across clients. We first formulate the problem of pruning LLMs in an FL setup and introduce a specialized aggregation function based on the $\ell_0$-norm to handle sparse model updates. Next, we propose an adaptive mask expansion technique to ensure that the global model meets the target sparsity, even when clients generate diverse pruning masks. Finally, we introduce a layer sampling strategy that allows clients to prune subsets of model layers based on their computational resources, enabling personalized pruning and reducing communication costs.

Figure~\ref{fig: framework} provides a visual overview of the FedSpaLLM framework, illustrating how clients and the central server interact during the pruning process.

\subsection{Problem Formulation}

We present the first formulation of pruning LLMs under the FL setting. In this scenario, multiple clients collaboratively prune a global LLM while ensuring that their local data remains private. Let $W_g$ denote the global model parameters, where $W_g \in \mathbb{R}^d$, and each client $i$ holds its own local dataset $\mathcal{D}_i = \{X_i, Y_i\}$ for training and pruning purposes.

During pruning, each client applies a binary pruning mask, $M_i \in \{0,1\}^d$, to its local model $W_i$. This mask determines which weights are retained ($M_i = 1$) and which are pruned ($M_i = 0$). The objective is to prune the global model while ensuring that the pruned models on each client still perform effectively on their respective local datasets. Importantly, our formulation allows for model heterogeneity or personalization, meaning that each client can have its own set of model parameters, $W_i$, which differs from the global model, $W_g$. This flexibility contrasts with the traditional FL setting, where all clients share the same global model.

The overall objective of federated pruning is to minimize the weighted sum of local losses across clients. Each client $i$ minimizes its local loss function $\mathcal{L}_i$ based on the pruned model $M_i \odot W_i$. Formally, this can be written as:
\begin{equation}
\min_{W_1, \ldots, W_N, M_1, \ldots, M_N} \sum\nolimits_{i=1}^{N} \alpha_i \cdot \mathcal{L}_i(M_i \odot W_i),
\end{equation}
subject to the global model $W_g$ being an aggregation of the locally pruned models:
\begin{equation}
W_g = \Omega\Big(M_1 \odot W_1, \ldots, M_N \odot W_N\Big),
\end{equation}
where $\Omega$ denotes the aggregation function that combines the pruned models from all clients to update the global model.

Our formulation is general and supports model heterogeneity, where each client may have its own model parameters, $W_i$, as opposed to the vanilla FL setting sharing a common global model. This flexibility is crucial for accommodating diverse client environments, making our method applicable to a wide range of real-world federated pruning scenarios.

\begin{figure*}[t!]
  \begin{center}
    \includegraphics[width=1\textwidth]{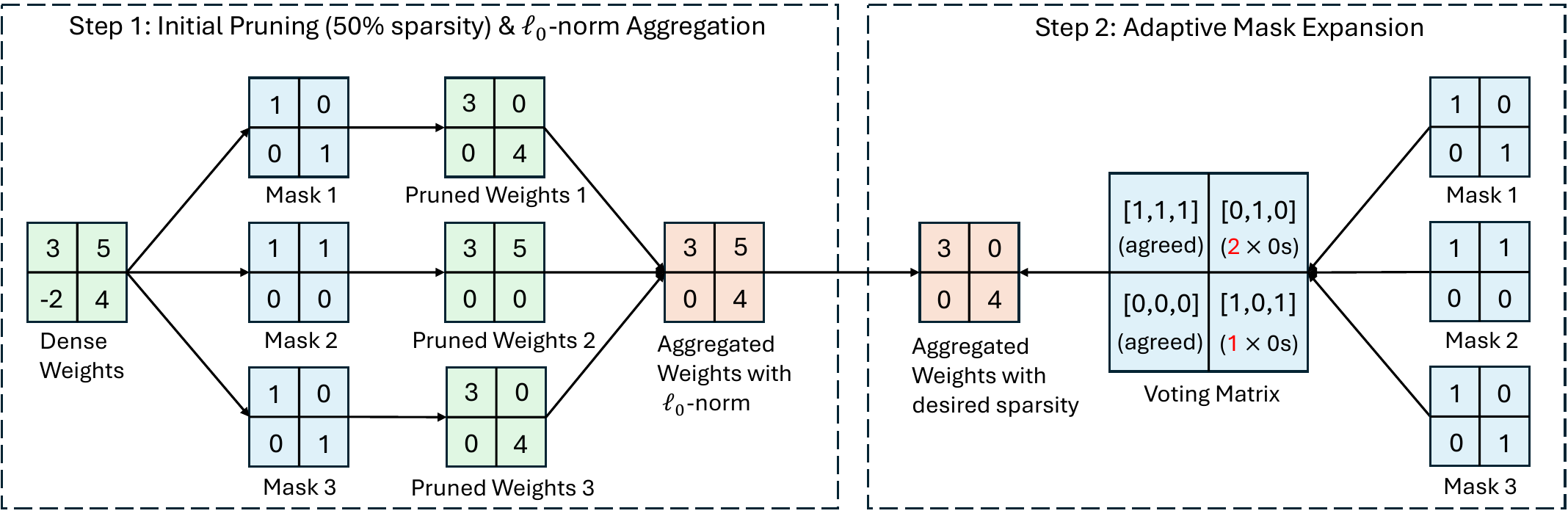}
  \end{center}
  \vspace{-2mm}
  \caption{Visualization of the proposed \textbf{Aggregation Function} of FedSpaLLM to handle heterogeneous sparsified parameters. After clients prune their local models, the server aggregates the pruned layers by using the $\ell_0$-norm aggregation function. This method avoids diluting the effect of unpruned weights by excluding zeros from the averaging process, thus preserving important parameters. To achieve the target global sparsity, an adaptive mask expansion is applied: the server counts the number of times each weight has been pruned across clients and uses this information to expand the pruning mask. The mask expansion prioritizes pruning weights that are most commonly pruned across clients, balancing individual client pruning decisions with the global sparsity goal.}
  \label{fig: aggregation}
  \vspace{-4mm}
\end{figure*}

\subsection{Aggregation Function}

\noindent\textbf{$\ell_{0}$-norm Aggregator.}
In traditional FL, the model weights from different clients are aggregated using a simple mean average or a similar function after each round of local updates. However, in our case, the parameters are sparsified through pruning, resulting in binary pruning masks. Due to the discrete nature of these masks, using a vanilla average function is not suitable. Specifically, when aggregating binary masks, dividing by the total count of clients would incorrectly include zeros (pruned positions) in the calculation, causing the averaged value to approach zero. This leads to an undesirable convergence, where the model parameters tend to vanish as the denominator grows larger than it should be. 

To address this, we propose an aggregation function based on the \(\ell_0\)-norm. Instead of averaging over all elements, we compute the average only over the non-zero elements (i.e., weights that are retained across clients). Mathematically, given the pruning masks $M_1, M_2, \ldots, M_N$ from $N$ clients and the corresponding local weights $W_1, W_2, \ldots, W_N$, the aggregation is defined as:
\begin{equation}
W_g = \frac{\sum_{i=1}^{N} M_i \odot W_i}{\left\| \sum_{i=1}^{N} M_i \right\|_{\ell_0}}.
\end{equation}
In this way, the aggregation only considers the non-zero elements, ensuring that the resulting global weights do not tend toward zero unless they are truly pruned across all clients. This method ensures the correct balance between sparsity and preserving important weights in the global model.

\begin{algorithm*}[t!]
\caption{FedSpaLLM: Federated Pruning of Large Language Models}
\begin{algorithmic}[1]
\Require Global LLM model $W_g$, Number of communication rounds $T$, Number of clients $N$, Sparsity target $s$, Client computational resources $\{r_i\}_{i=1}^N$
\Ensure Pruned global model $W_g'$
\State Initialize global model $W_g$
\For{each communication round $t = 1$ to $T$}
    \State \textbf{Server:} Sample a subset of layers for each client based on computational resources
    \For{each client $i = 1$ to $N$}
        \State Sample $k_i$ layers: $L_i \gets \text{Sample}(L, k_i), \text{where}~ k_i \propto r_i$
        \State Send only the sampled layers $L_i$ to client $i$
    \EndFor
    \For{each client $i = 1$ to $N$ \textbf{in parallel}}
        \State \textbf{Client:} Prune the received layers $L_i$ with local data
        \State Use pre-cached dense layers for unsampled layers
        \State Generate pruning mask $M_i$ for layers $L_i$: $M_i \gets \text{GeneratePruningMask}(W_g[L_i], \text{LocalData}_i)$
        \State Prune and update local model: $W_i' \gets M_i \odot W_g[L_i]$
        \State Send pruned weights $W_i'[L_i]$ and mask $M_i$ back to the server
    \EndFor
    \State \textbf{Server:} Aggregate pruned layers using $\ell_0$-norm aggregation
    \State $W_g' \gets \ell_0\text{-NormAggregation}(\{W_i'[L_i], M_i\}_{i=1}^N)$
    \State Apply adaptive mask expansion to achieve target sparsity $s$
    \State Adjust global pruning mask to meet desired sparsity level
\EndFor
\State \Return Final pruned global model $W_g'$
\end{algorithmic}
\end{algorithm*}

\noindent\textbf{Adaptive Aggregation w/. Mask Expansion.}
Another challenge arises from the heterogeneity of data across clients. Even with the same target sparsity, different clients may generate different pruning masks based on their local data. If we simply averaged these masks, the result would be equivalent to taking the intersection of all local masks. In other words, only the weights pruned by all clients would be pruned in the global model. This leads to a situation where the global model's sparsity is always smaller than the target sparsity, as fewer weights are pruned overall.

To address this, we propose an \emph{mask expansion} technique. The idea is to first apply the \(\ell_0\)-norm aggregation described above and then expand the global pruning mask to achieve the target sparsity. Specifically, after the initial aggregation, we count the number of zeros (pruned positions) for each weight across clients. If the count is below the target sparsity, we sort the remaining weights and select the ones with the most agreement (i.e., those that are most commonly pruned) to prune further, ensuring the final sparsity matches the desired level.

Let $C_j$ represent the number of zeros for the $j$-th weight across clients, where:
\begin{equation}
C_j = \sum\nolimits_{i=1}^{N} \mathbb{I}(M_i^{(j)} = 0),
\end{equation}
where $\mathbb{I}(\cdot)$ is the indicator function that returns 1 if $M_i^{(j)} = 0$ (the $j$-th weight is pruned by client $i$) and 0 otherwise. After counting, we sort the weights based on their $C_j$ values and expand the pruning mask for weights that are not fully agreed upon by all clients. To achieve the target sparsity $s$, we select the top-$k$ weights from the sorted list, where:
\begin{equation}
k = \text{ceil}(s \cdot d) - \sum\nolimits_{j=1}^{d} \mathbb{I}(C_j = N),
\end{equation}
where $d$ is the total number of weights and $s$ is the target sparsity. The first term ensures that the final mask achieves the desired sparsity, and the second term subtracts the number of weights that are already pruned by all clients (i.e., those where $C_j = N$).

By employing this sorting and expansion method, we ensure that the global pruning mask adheres to the desired sparsity while reflecting the commonly agreed-upon pruned positions, balancing individual client decisions and the global sparsity requirement.

\subsection{Personalization with Layer Sampling}

In FL for LLM pruning, communication overhead, and system heterogeneity are key challenges. The pruning of LLMs is typically done in a local manner, where the pruning of each decoder layer is independent, allowing for efficient and flexible layer sampling. This independence enables us to design a novel sampling strategy that maintains both efficiency and pruning accuracy while addressing the diverse computational capacities of clients. 

\noindent\textbf{Remark:} Unlike traditional FL, where models are typically fine-tuned holistically, pruning in LLMs allows each decoder layer to be pruned independently. We exploit this property by sampling layers, significantly reducing communication costs without compromising the overall pruning outcome.

\noindent\textbf{Layer Sampling Strategy.}
In each communication round, the server randomly samples a subset of layers from the LLM for each client to prune. Let $L = \{l_1, l_2, \dots, l_m\}$ denote the set of all layers in the LLM, where $m$ is the total number of layers. For each client $i$, the server selects a subset of layers $L_i \subseteq L$ to be pruned, where the number of layers sampled, $|L_i| = k_i$, is proportional to the computational capacity of client $i$, denoted as $r_i$. Formally, 
\begin{equation}
L_i = \text{Sample}(L, k_i), \quad k_i \propto r_i.
\end{equation}

Once client $i$ receives the subset $L_i$, it performs pruning on the sampled layers while retaining the original weights of the unsampled layers. Let $W_g = \{W_1, W_2, \dots, W_m\}$ represent the global model weights, and $M_i = \{M_1, M_2, \dots, M_m\}$ represent the pruning masks for client $i$. The locally pruned model $W_i^\prime$ is updated as:
\begin{equation}
W_i^\prime = 
\begin{cases} 
      M_j \odot W_j, & \text{if }  j \in L_i, \\
      W_{j}^{\text{dense}}, & \text{if } j \notin L_i.
   \end{cases}
\end{equation}
For unsampled layers $j \notin L_i$ we retain their original unpruned dense weights $W_j^{\text{dense}}$. This process ensures that communication costs are minimized, 
% as only a portion of the model is transmitted during each round. \kk{What is communicated?}
as we only need to communicate the sampled layers $L_i$, only a portion of the model, in each round. 

\begin{corollary}[Sparsity Guarantee]
Let $\mathcal{S}_{global}$ denote the target global sparsity, and let $\mathcal{S}_i$ be the sparsity achieved by client $i$ on its local model. If the layer sampling strategy ensures that all layers $\mathcal{L}$ are sampled at least once across all clients in each communication round, the aggregated global pruned model $\hat{W}_g$ will maintain the same sparsity as the target sparsity $\mathcal{S}_{global}$. Since all clients share the same target sparsity, and our $\ell_0$-norm aggregation guarantees that the aggregated layers have the exact sparsity as the locally pruned layers, the sparsity of the global model will always equal the desired target. Formally:
\begin{equation}
\mathcal{S}_{global} = \mathcal{S}_i, \quad \forall i,
\end{equation}
where $N$ is the total number of clients. Thus, the sparsity of the global model is consistent with the target across all communication rounds.
\end{corollary}

This corollary ensures that, with our sampling strategy, the global model will meet the desired global sparsity after aggregation. 

\begin{theorem}[Unbiased Estimator]
Let $\hat{W}_g$ denote the global model obtained by aggregating pruned models after layer sampling, and $W_g^*$ be the global model obtained if all layers were pruned at every client (without sampling). The model $\hat{W}_g$ is an unbiased estimator of $W_g^*$ under the layer sampling strategy. Formally:
\begin{equation}
\mathbb{E}[\hat{W}_g] = W_g^*.
\end{equation}
This holds as long as each layer has an equal probability of being selected across clients during each communication round, ensuring that all layers are eventually represented in the final aggregated model.
\end{theorem}

Theorem 1 ensures that the global model obtained by layer sampling converges to the same result as if every client had pruned all layers. The key here is the randomness of the layer selection process: over multiple communication rounds, the expected contribution of each layer is preserved, resulting in an unbiased estimate of the fully pruned model. This guarantees that our sampling strategy does not introduce systematic errors and that, over time, the sampled model will mirror the performance of the fully pruned model.

\begin{table*}[!th] 
  \scriptsize
  \centering
  \begin{tabular}{p{1.2cm}p{0.8cm}p{0.8cm}p{0.8cm}p{0.8cm}p{0.8cm}p{0.8cm}p{0.8cm}p{0.8cm}p{0.8cm}p{0.8cm}p{0.8cm}p{0.8cm}}
    \toprule
    \multicolumn{13}{c}{OPT-125m} \\
    \midrule
    Sparsity & \multicolumn{3}{c}{50\%} & \multicolumn{3}{c}{60\%} & \multicolumn{3}{c}{70\%} & \multicolumn{3}{c}{80\%} \\
    \cmidrule(lr){2-4} \cmidrule(lr){5-7} \cmidrule(lr){8-10} \cmidrule(lr){11-13}
    Dataset & WT-2 & PTB & C4 & WT-2 & PTB & C4 & WT-2 & PTB & C4 & WT-2 & PTB & C4\\
    \midrule
    Random & 1.45e4 & 1.36e4 & 1.17e4 & 1.86e4 & 1.74e4 & 1.62e4 & 2.16e4 & 2.09e4 & 1.96e4 & 3.22e4 & 3.22e4 & 3.03e4 \\
    Standalone & 37.87 & 57.38 & 33.78 & 62.09 & 93.14 & 49.38 & 237.07 & 296.61 & 158.54 & 1699.62 & 1907.27 &  759.18 \\
    \cellcolor{green}FedSpaLLM & 37.65 & 57.07 & 33.75 & 61.28 & 92.32 & 48.96 & 226.44 & 287.72 & 152.96 & 1414.77 & 1699.16 & 739.97\\
 
    \toprule
    \multicolumn{13}{c}{OPT-1.3b} \\
    \midrule
    Sparsity & \multicolumn{3}{c}{50\%} & \multicolumn{3}{c}{60\%} & \multicolumn{3}{c}{70\%} & \multicolumn{3}{c}{80\%} \\
    \cmidrule(lr){2-4} \cmidrule(lr){5-7} \cmidrule(lr){8-10} \cmidrule(lr){11-13}
    Dataset & WT-2 & PTB & C4 & WT-2 & PTB & C4 & WT-2 & PTB & C4 & WT-2 & PTB & C4\\
    \midrule
    Random & 1.90e4 & 1.92e4 & 1.75e4 & 1.77e4 & 1.73e4 & 1.67e4 & 2.14e4 &	2.23e4 & 2.01e4 & 3.00e4 & 3.10e4 & 2.97e4 \\
    Standalone & 18.16 & 26.78 & 20.31 & 23.55 & 35.93 & 24.15 & 62.08 & 97.46 & 52.13 & 1814.44 & 1756.18 & 636.56\\
    \cellcolor{green}FedSpaLLM & 18.09 & 26.52 & 20.13 & 22.70 & 34.05 & 23.72 & 54.87 & 82.99 & 48.05 & 1653.65 & 1564.86 & 598.83 \\
   
    \toprule
    \multicolumn{13}{c}{LlaMA-2 7b} \\
    \midrule
    Sparsity & \multicolumn{3}{c}{50\%} & \multicolumn{3}{c}{60\%} & \multicolumn{3}{c}{70\%} & \multicolumn{3}{c}{80\%} \\
    \cmidrule(lr){2-4} \cmidrule(lr){5-7} \cmidrule(lr){8-10} \cmidrule(lr){11-13}
    Dataset & WT-2 & PTB & C4 & WT-2 & PTB & C4 & WT-2 & PTB & C4 & WT-2 & PTB & C4\\
    \midrule
    Random & 8.83e4 & 1.20e5 & 1.41e5  & 1.98e5 & 2.99e5 & 1.92e5 & 5.06e4 & 4.98e4 & 4.93e4 & 4.99e4 & 4.91e4 & 4.97e4 \\
    Standalone & 7.13 & 185.18 & 9.47 & 10.71 & 1255.83 & 13.52 & 31.60 & 6582.79 & 34.17 & 124.59 & 8398.51 & 110.45 \\
    \cellcolor{green}FedSpaLLM & 6.71 & 88.15 & 9.03 & 9.02 & 300.35 & 11.94 & 20.14 & 1371.44 & 23.28 & 119.21 & 3382.99 & 98.87\\
    \bottomrule
  \end{tabular}
  \vspace{-3mm}
  \caption{Average perplexity of the client models and perplexity of the global model with 4 clients; the lower the perplexity, the better.} \label{tab:results_4clients}
  \vspace{-2mm}
\end{table*}

\begin{table*}[!th] 
  \scriptsize
  \centering
  \begin{tabular}{p{1.2cm}p{0.8cm}p{0.8cm}p{0.8cm}p{0.8cm}p{0.8cm}p{0.8cm}p{0.8cm}p{0.8cm}p{0.8cm}p{0.8cm}p{0.8cm}p{0.8cm}}
    \toprule
    \multicolumn{13}{c}{OPT-125m} \\
    \midrule
    Sparsity & \multicolumn{3}{c}{50\%} & \multicolumn{3}{c}{60\%} & \multicolumn{3}{c}{70\%} & \multicolumn{3}{c}{80\%} \\
    \cmidrule(lr){2-4} \cmidrule(lr){5-7} \cmidrule(lr){8-10} \cmidrule(lr){11-13}
    Dataset & WT-2 & PTB & C4 & WT-2 & PTB & C4 & WT-2 & PTB & C4 & WT-2 & PTB & C4\\
    \midrule
    Random & 1.24e4 & 1.12e4  & 9.93e3 & 1.65e4 & 1.52e4 & 1.40e4 & 2.11e4 & 2.04e4 & 1.86e4 & 3.29e4 & 3.31e4 & 3.15e4 \\
    Standalone & 38.40 & 57.51 & 33.95 & 64.52 & 94.19 & 50.08 & 244.90 & 308.59 & 158.49 & 1655.23 & 1893.9 & 732.29 \\
    \cellcolor{green}FedSpaLLM & 38.01 & 56.66 & 33.65 & 63.00 & 91.50 & 49.10 & 217.54 & 274.06 & 147.58 & 1415.78 & 1611.45 & 665.36\\
 
    \toprule
    \multicolumn{13}{c}{OPT-1.3b} \\
    \midrule
    Sparsity & \multicolumn{3}{c}{50\%} & \multicolumn{3}{c}{60\%} & \multicolumn{3}{c}{70\%} & \multicolumn{3}{c}{80\%} \\
    \cmidrule(lr){2-4} \cmidrule(lr){5-7} \cmidrule(lr){8-10} \cmidrule(lr){11-13}
    Dataset & WT-2 & PTB & C4 & WT-2 & PTB & C4 & WT-2 & PTB & C4 & WT-2 & PTB & C4\\
    \midrule
    Random & 1.59e4	& 1.44e4 & 1.40e4 & 2.15e4 & 2.17e4 & 1.84e4 & 2.26e4 & 2.27e4 & 2.06e4	& 3.25e4 & 3.27e4 & 3.22e4 \\
    Standalone & 18.51 & 27.18 & 20.29 & 23.99 & 36.53 & 24.54 & 68.70 & 108.93 & 56.28 & 2109.46 & 2134.22 & 733.80 \\
    \cellcolor{green}FedSpaLLM & 18.38 & 27.89 & 20.03 & 23.20 & 35.60 & 23.75 & 63.31 & 105.37 & 54.85 & 1979.77 & 1989.21 & 690.86\\
   
    \toprule
    \multicolumn{13}{c}{LlaMA-2 7b} \\
    \midrule
    Sparsity & \multicolumn{3}{c}{50\%} & \multicolumn{3}{c}{60\%} & \multicolumn{3}{c}{70\%} & \multicolumn{3}{c}{80\%} \\
    \cmidrule(lr){2-4} \cmidrule(lr){5-7} \cmidrule(lr){8-10} \cmidrule(lr){11-13}
    Dataset & WT-2 & PTB & C4 & WT-2 & PTB & C4 & WT-2 & PTB & C4 & WT-2 & PTB & C4\\
    \midrule
    Random & 2.62e5 & 1.14e5 & 2.92e5 & 6.57e4 & 6.11e4 & 6.58e4 & 6.15e4 & 6.25e4 & 6.17e4 & 4.57e4 & 4.76e4 & 4.78e4 \\
    Standalone & 7.17 & 193.89 & 9.43 & 10.76 & 1126.63 & 13.45 & 30.77 & 8349.21 & 33.31 & 134.08 & 1.3e4 & 108.87 \\
    \cellcolor{green}FedSpaLLM & 6.70 & 73.67 & 8.97 & 9.09 & 175.98 & 12.03 & 21.18 & 1242.25 & 24.98 & 117.73 & 2819.17 & 96.44\\
    \bottomrule
  \end{tabular}
  \vspace{-3mm}
  \caption{Average perplexity of the client models and perplexity of the global model with 8 clients; the lower the perplexity, the better.} \label{tab:results_8clients}
  \vspace{-2mm}
\end{table*}

\begin{table*}[!th] 
  \scriptsize
  \centering
  \begin{tabular}{p{1.2cm}p{0.8cm}p{0.8cm}p{0.8cm}p{0.8cm}p{0.8cm}p{0.8cm}p{0.8cm}p{0.8cm}p{0.8cm}p{0.8cm}p{0.8cm}p{0.8cm}}
    \toprule
    \multicolumn{13}{c}{OPT-125m} \\
    \midrule
    Sparsity & \multicolumn{3}{c}{50\%} & \multicolumn{3}{c}{60\%} & \multicolumn{3}{c}{70\%} & \multicolumn{3}{c}{80\%} \\
    \cmidrule(lr){2-4} \cmidrule(lr){5-7} \cmidrule(lr){8-10} \cmidrule(lr){11-13}
    Dataset & WT-2 & PTB & C4 & WT-2 & PTB & C4 & WT-2 & PTB & C4 & WT-2 & PTB & C4\\
    \midrule
    Random & 1.22e4 & 1.14e4 & 9.96e3 & 1.66e4 & 1.56e4 & 1.45e4 & 2.47e4 & 2.35e4 & 2.24e4 	& 3.20e4 & 3.23e4 & 2.95e4 \\
    Standalone & 38.24 & 57.37 & 33.91 & 67.42 & 98.29 & 51.92 & 264.88 & 326.87 & 171.83 & 1624.46 & 1764.94 & 777.79\\
    \cellcolor{green}FedSpaLLM & 38.00 & 56.90 & 33.62 & 66.41 & 97.92 & 51.84 & 240.18 & 304.17 & 167.30 & 1389.22 & 1454.49 & 700.84\\
 
    \toprule
    \multicolumn{13}{c}{OPT-1.3b} \\
    \midrule
    Sparsity & \multicolumn{3}{c}{50\%} & \multicolumn{3}{c}{60\%} & \multicolumn{3}{c}{70\%} & \multicolumn{3}{c}{80\%} \\
    \cmidrule(lr){2-4} \cmidrule(lr){5-7} \cmidrule(lr){8-10} \cmidrule(lr){11-13}
    Dataset & WT-2 & PTB & C4 & WT-2 & PTB & C4 & WT-2 & PTB & C4 & WT-2 & PTB & C4\\
    \midrule
    Random & 1.80e4 & 1.77e4 & 1.53e4 & 1.88e4 & 1.87e4 & 1.75e4 & 2.32e4 & 2.22e4 & 2.14e4	& 2.99e4 & 3.02e4	& 2.94e4 \\
    Standalone & 18.99 & 27.48 & 20.76 & 25.01 & 37.87 & 25.49 & 82.24 & 138.91 & 65.53 & 2472.21 & 2475.89 & 873.45 \\
    \cellcolor{green}FedSpaLLM & 18.20 & 27.30 & 20.52 & 23.93 & 36.36 & 22.72 & 77.51 & 129.40 & 63.62 & 2327.93 & 2361.86 & 838.53\\
   
    \toprule
    \multicolumn{13}{c}{LlaMA-2 7b} \\
    \midrule
    Sparsity & \multicolumn{3}{c}{50\%} & \multicolumn{3}{c}{60\%} & \multicolumn{3}{c}{70\%} & \multicolumn{3}{c}{80\%} \\
    \cmidrule(lr){2-4} \cmidrule(lr){5-7} \cmidrule(lr){8-10} \cmidrule(lr){11-13}
    Dataset & WT-2 & PTB & C4 & WT-2 & PTB & C4 & WT-2 & PTB & C4 & WT-2 & PTB & C4\\
    \midrule
    Random & 1.76e5 & 2.79e5 & 2.19e5 & 8.30e4 & 7.96e4 & 7.92e4 & 9.73e4 & 2.54e5 & 1.13e5 & 5.20e4 & 4.89e4 & 4.89e4 \\
    Standalone & 7.17 & 188.35 & 9.41 & 10.80 & 969.84 & 13.44 & 31.81 & 6671.71 & 33.20 & 137.21 & 7757.43 & 106.87\\
    \cellcolor{green}FedSpaLLM & 6.75 & 70.79 & 9.04 & 9.33 & 178.15 & 12.45 & 25.09 & 702.03 & 27.27 & 115.91 & 1791.83 & 94.87\\
    \bottomrule
  \end{tabular}
  \vspace{-3mm}
  \caption{Average perplexity of the client models and perplexity of the global model with 16 clients; the lower the perplexity, the better.} \label{tab:results_16clients}
  \vspace{-4mm}
\end{table*}

\section{Experiments}
\label{sec:experiments}

\paragraph{Experiments Setup.}
We implement our FedSpaLLM in PyTorch~\cite{paszke2019pytorch} and use the HuggingFace Transformers library~\cite{wolf2019huggingface} for handling models and datasets. All pruning experiments are conducted on NVIDIA A100 GPUs. For each client, we utilize SparseGPT~\cite{frantar2023massive} to perform pruning. For the calibration data, we follow \cite{frantar2023massive} and use 2048-token segments, randomly chosen from the first shard of the C4~\cite{raffel2020exploring} dataset. This represents generic text data crawled from the internet and ensures that our experiments remain zero-shot since no task-specific data is seen during pruning. In addition, we consider a random pruning baseline in which the model is randomly pruned to the target sparsity.

In particular, we perform experiments on the OPT model family~\cite{zhang2022opt} and LlaMA-2 model family~\cite{touvron2023llama} with 4, 8, and 16 clients. We consider OPT-125m, OPT-1.3b, and LlaMA-2 7b with unstructured sparsity of 50\% to 80\%. In each communication round, each of the clients receives a copy of the same global model from the server and each client is assumed to utilize 32 calibration samples to perform its own pruning. By evaluating models of varying sizes alongside different number of clients, we can gain a more comprehensive understanding of FedSpaLLM's performances in its scalability and robustness. In terms of metrics, we mainly focus on perplexity, which is known to be a challenging and stable metric that is well-suited for evaluating the accuracy of compression methods~\cite{yao2022zeroquant,dettmers2023case}, and thus measuring the performances of the compressed models. 
We consider the test sets of raw-WikiText2~\cite{merity2016pointer} (WT-2) and PTB~\cite{marcus1994penn} as well as a subset of the C4 validation data, which are all popular benchmarks in LLM compression literature~\cite{yao2022zeroquant,park2022nuqmm,frantar2023massive,sun2023simple}. 

\subsection{Results and analyses}
We present the perplexity results in Tables~\ref{tab:results_4clients} to \ref{tab:results_16clients}. In the tables, we report the random pruning baseline, denoted by ``Random", the average perplexity of the client models, denoted by ``standalone", and the global model, denoted by ``FedSpaLLM". From the results, we see that random pruning results in perplexity that are orders of magnitude higher than both standalone and FedSpaLLM. As the model size increases, the performances of random pruning becomes even worse. This suggests that random pruning may have pruned weights that are crucial for maintaining model quality. Comparing standalone and FedSpaLLM, we see that across the models, datasets, and sparsity levels, FedSpaLLM consistently outperforms standalone in achieving lower perplexity. In general, as the target sparsity increases, we see more noticeable improvements in the perplexity of the global model over the client models. This is expected because as the sparsity increases, more information is required to accurately prune the model weights to maintain the model performances. In essence, each of the client models is pruned with the private calibration samples of each client while the global model benefits from the collaborative information from the communicated weights from the clients. As a result, the global model is effectively utilizing a signicantly larger calibration sample set, even though such a centralized calibration sample set is prohibited in FL setting as the client's calibration samples are private. Notably, FedSpaLLM achieves substantially lower perplexity compared to standalone in higher sparsity levels, highlighting the benefits of FedSpaLLM where the clients collaboratively contribute to the global model with much better performances while the privacy of their own data is well maintained. Furthermore, we can see the improvements in the perplexity of the global model over the client models are particularly significant for the LlaMA-2 model family and the PTB dataset. We observe that, in the case of LlaMA-2 7b, the client models generally struggle with the PTB dataset from sparsity 60\% and beyond, regardless of the number of clients. In many of those cases, the global model achieves up to 5 times better perplexity values. This demonstrates the effectiveness of FL in the pruning tasks. In addition, we do not observe there is noticeable trend in perplexity values with varying number of clients and the perplexity values of the global models are comparable regardless of the number of clients participating in FL, when the sparsity level is small.

\section{Conclusion}
\label{sec:conclusion}

In this paper, we introduced FedSpaLLM, a novel framework for pruning LLMs in federated learning settings. By addressing key challenges such as communication efficiency and system heterogeneity, FedSpaLLM enables collaborative pruning without sharing private data. Our method introduces several innovations, including the use of $\ell_0$-norm aggregation for handling sparse model updates, an adaptive mask expansion technique to ensure target global sparsity and a layer sampling strategy that personalizes the pruning process based on client resources.

Through extensive experiments, we demonstrated that FedSpaLLM significantly reduces the communication and computational costs of LLM pruning while maintaining model performance across diverse, real-world federated environments. Our results showed that FedSpaLLM outperforms existing approaches in global model perplexity, making it a promising solution for resource-constrained applications where privacy is critical.

In future work, we aim to extend FedSpaLLM to other model compression techniques, such as quantization and distillation, to further improve efficiency in federated learning scenarios. Additionally, exploring more advanced techniques for client selection and resource-aware scheduling could enhance the adaptability and scalability of the framework to even larger LLMs and more heterogeneous environments.

\section{Limitations}

While FedSpaLLM introduces a highly efficient framework for pruning large language models in federated settings, there are a few limitations to consider. First, the effectiveness of the layer sampling strategy depends on the accurate estimation of client computational resources, which may vary dynamically in real-world deployments. This could lead to suboptimal layer sampling in highly fluctuating environments. Second, while our adaptive mask expansion ensures global sparsity, further refinements could improve its handling of extreme heterogeneity in client data distributions. Finally, our current experiments focus on moderate-scale LLMs, and additional work is required to assess scalability for larger models beyond several billion parameters.

\section{Acknowledgement}
This work was supported by the U.S. Department of Energy, Office of Science, Advanced Scientific Computing Research, under Contract DE-AC02-06CH11357.
This research used resources of the Argonne Leadership Computing Facility at Argonne National Laboratory, which is supported by the Office of Science of the U.S. Department of Energy under contract DE-AC02-06CH11357.

\bibliography{references}

\clearpage
\appendix

\section{Theoretical Proofs}

\subsection{Proof of Corollary 1 (Sparsity Guarantee)}

\begin{proof}
1. \textbf{Client Sparsity Consistency:}  
Each client enforces the same target sparsity $\mathcal{S}_{global}$. This implies that for each client $i$, the sparsity of the pruned layers $\mathcal{L}_i$ matches the global sparsity target $\mathcal{S}_{global}$. Formally, we have:
\[
\mathcal{S}_i = \mathcal{S}_{global}, \quad \forall i = 1, 2, \dots, N.
\]
Since all clients prune their local models independently but according to the same target sparsity, each pruned local model achieves the same sparsity.

2. \textbf{Layer Sampling Strategy:}  
The layer sampling strategy ensures that all layers of the model $\mathcal{L}$ are eventually sampled across all clients during each communication round. Therefore, every layer in the global model $\hat{W}_g$ has been pruned by each client according to the same sparsity criterion $\mathcal{S}_{global}$.

3. \textbf{Aggregation with $\ell_0$-norm:}  
The aggregation function using $\ell_0$-norm averages only the non-zero elements (i.e., the pruned weights) from the client models. Since all clients enforce the same sparsity, and the aggregation only involves the non-zero weights from these pruned models, the sparsity of the aggregated global model will match that of the local models. Specifically:
\[
\mathcal{S}_{global} = \mathcal{S}_i, \quad \forall i.
\]
Thus, the sparsity of the global model after aggregation is equivalent to the sparsity of each client model.

4. \textbf{Conclusion:}  
Therefore, the global model $\hat{W}_g$ will maintain the target sparsity $\mathcal{S}_{global}$ after aggregation in each communication round. The aggregation process ensures that the global sparsity is consistent with the target sparsity across rounds.

\[
\boxed{\mathcal{S}_{global} = \mathcal{S}_i, \quad \forall i.}
\]

\end{proof}

\subsection{Proof of Theorem 1 (Unbiased Estimator)}

\begin{proof}
1. \textbf{Layer Sampling Strategy:}  
Let $\mathcal{L} = \{L_1, L_2, \dots, L_m\}$ denote the set of all layers in the model, where $m$ is the total number of layers. In each communication round, the server randomly samples a subset of layers $\mathcal{L}_i \subseteq \mathcal{L}$ for each client $i$. Each layer $L_j \in \mathcal{L}$ has an equal probability $p_j$ of being selected across clients.

The expectation of the sampled weights for layer $L_j$ across all clients can be expressed as:
\[
\mathbb{E}[W_i[L_j]] = p_j W_g^*[L_j],
\]
where $W_g^*[L_j]$ is the weight of layer $L_j$ in the fully pruned global model $W_g^*$.

2. \textbf{Unbiasedness of Layer Sampling:}  
Since each layer has an equal probability of being sampled across clients, the expected contribution of each layer is proportional to its selection probability. Over multiple communication rounds, all layers will be sampled enough times to represent the fully pruned model.

Therefore, the expected value of the global model $\hat{W}_g$ is the same as the fully pruned model $W_g^*$. For any layer $L_j$, we have:
\begin{equation}
    \begin{split}
        \mathbb{E}[\hat{W}_g[L_j]] &= \mathbb{E}\left[\frac{1}{N} \sum_{i=1}^N W_i[L_j]\right] \\
        &= \frac{1}{N} \sum_{i=1}^N \mathbb{E}[W_i[L_j]] = W_g^*[L_j].
    \end{split}
\end{equation}
Thus, the global model $\hat{W}_g$ is an unbiased estimator of $W_g^*$, as the expected value of the pruned weights matches the fully pruned model.

3. \textbf{Conclusion:}  
Therefore, the global model $\hat{W}_g$ obtained by aggregating pruned models after layer sampling is an unbiased estimator of the fully pruned model $W_g^*$. Formally, we can conclude that:
\[
\boxed{\mathbb{E}[\hat{W}_g] = W_g^*}.
\]
This unbiased property holds as long as each layer has an equal probability of being selected across clients during each communication round.
\end{proof}

\end{document}